\documentclass[lettersize,journal]{IEEEtran}
\usepackage{amsmath,amsfonts}
\usepackage{algorithmic}
\usepackage{algorithm}
\usepackage{array}
\usepackage{textcomp}
\usepackage{stfloats}
\usepackage{url}
\usepackage{verbatim}
\usepackage{graphicx}
\usepackage{cite}
\usepackage{lipsum}
\usepackage{multirow}
\usepackage{amsmath}
\usepackage[table,xcdraw]{xcolor}
\usepackage{pifont}
\usepackage{dsfont}
\usepackage{booktabs}
\usepackage{float}
\usepackage{caption}
\usepackage{subcaption}
\usepackage{tabularx}
\usepackage{array}

\usepackage{hyperref}

\definecolor{MyDarkGreen}{HTML}{008585}
\definecolor{MyLightGreen}{HTML}{74a892}
\definecolor{MyDarkRed}{HTML}{c7522a}
\definecolor{MyDarkYellow}{HTML}{e5c185}

\begin{document}

\title{Rethinking Metrics and Benchmarks of Video Anomaly Detection}

\author{Zihao Liu, Xiaoyu Wu, Wenna Li, Linlin Yang, Shengjin Wang, ~\IEEEmembership{Senior Member,~IEEE,}
\thanks{Zihao Liu, Xiaoyu Wu, Wenna Li, Linlin Yang are with the State Key Laboratory of Media Convergence and Communication, Communication University
of China, Beijing, China (e-mail: liuzihao@cuc.edu.cn; wuxiaoyu@cuc.edu.cn; liwenna@cuc.edu.cn; mu4yang@gmail.com). Shengjin Wang is with the Department of Electronic Engineering, Tsinghua University, Beijing, China (e-mail: wgsgj@tsinghua.edu.cn)}%
\thanks{Corresponding author: Xiaoyu Wu.}
\thanks{Manuscript received April 19, 2021; revised August 16, 2021.}}

\markboth{Journal of \LaTeX\ Class Files,~Vol.~14, No.~8, August~2021}%
{Shell \MakeLowercase{\textit{et al.}}: A Sample Article Using IEEEtran.cls for IEEE Journals}

\IEEEpubid{0000--0000/00\$00.00~\copyright~2021 IEEE}

\maketitle

\begin{abstract}
Video Anomaly Detection (VAD), which aims to detect anomalies that deviate from expectation, has attracted increasing attention in recent years.
Existing advancements in VAD primarily focus on model architectures and training strategies, while devoting insufficient attention to evaluation metrics and benchmarks. 
In this paper, we rethink VAD evaluation methods through comprehensive analyses, revealing three critical limitations in current practices: 1) existing metrics are significantly influenced by single annotation bias; 2) current metrics fail to reward early detection of anomalies; 3) available benchmarks lack the capability to evaluate scene overfitting of fully/weakly-supervised algorithms.
To address these limitations, we propose three novel evaluation methods: first, we establish probabilistic AUC/AP (Prob-AUC/AP) metrics utlizing multi-round annotations to mitigate single annotation bias; second, we develop a Latency-aware Average Precision (LaAP) metric that rewards early and accurate anomaly detection; and finally, we introduce two hard normal benchmarks (UCF-HN, MSAD-HN) with videos specifically designed to evaluate scene overfitting.
We report performance comparisons of ten state-of-the-art VAD approaches using our proposed evaluation methods, providing novel perspectives for future VAD model development.
We release our data and code in \url{https://github.com/Kamino666/RethinkingVAD}.
\end{abstract}


\section{Introduction}

Video Anomaly Detection (VAD) aims to identify abnormal frames deviating from expectations in video sequences, playing a pivotal role in applications such as intelligent surveillance, traffic management, and healthcare \cite{VAD-review-2023,VAD-review-wupeng}. In recent years, this task has garnered increasing attention, with existing methods \cite{RTFM,MGFN,URDMU,UMIL,TEVAD,CLIP-TSA,PEL,vadclip,NormalityGuidance,NormalityGuided,fan2024tcsvt,yang2023tcsvt,xu2025plovad} achieving satisfactory performance on several large-scale datasets \cite{ucf-crime,msad,xdviolence}. 
Similar to other domains in machine learning, the advancement of VAD correlates with evaluation metrics and datasets. Despite its importance, this direction remains understudied. Inspired by similar studies in other domains \cite{CloserLookVMR,UncoveringVMR}, we conduct a comprehensive analysis of the methods of evaluating VAD algorithms. 
The scope of this study is delineated as the temporal localization of anomalies in videos under general and unconstrained scenes.
Our investigation reveals that existing metrics and benchmarks fail to reliably and comprehensively evaluate VAD algorithms. 
We summarize the limitations in terms of annotation, metric, and benchmarking data as follows.

\begin{figure*}[t]
    \centering
    \includegraphics[width=0.95\linewidth]{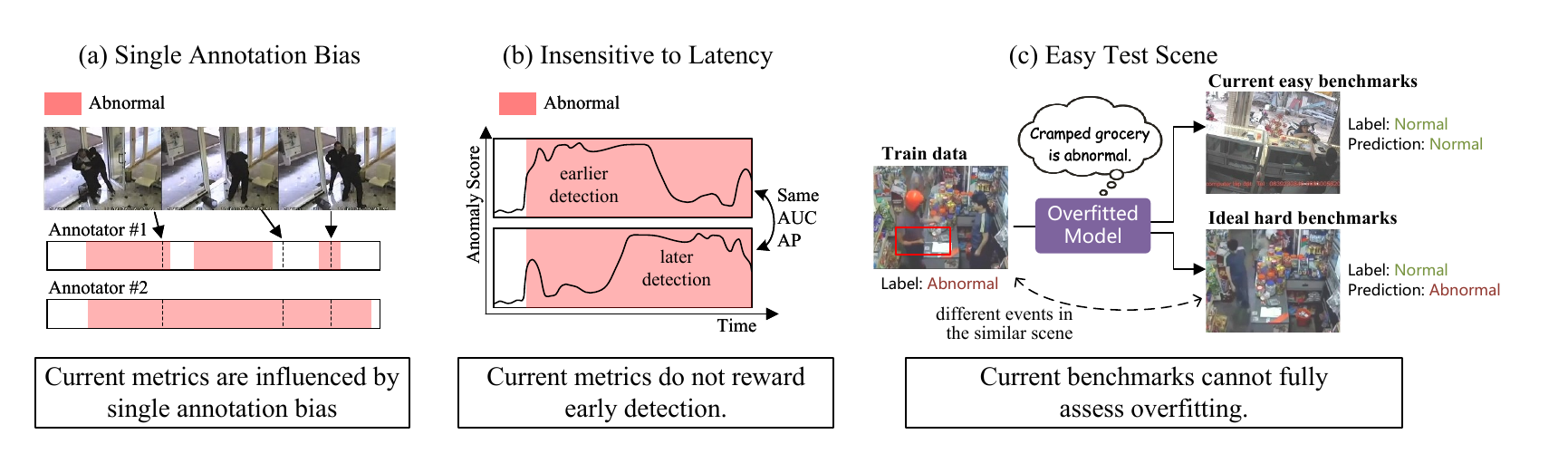}
    \caption{Three limitations in current metrics and benchmarks of video anomaly detection.}
    \label{fig:motivation}
    \vspace{-1em}
\end{figure*}

For annotation, existing evaluation methods are generally limited to leveraging single-round annotations, where each video has only single hard label. However, individual annotations may exhibit bias. In Figure~\ref{fig:motivation}(a), one tends to favor the most salient abnormal behavior frames while the other may favor more comprehensive event segments. The discrepancy of annotations could compromise the reliability of metric computations.
\cite{ECCV_CrossDomain_LimitedSup} preliminarily reveals this phenomenon in the UCF-Crime dataset \cite{ucf-crime} by providing an alternative annotation. Some works \cite{RealTimeWSVAD,UCA} also mention the label noise in UCF-Crime. However, these works lack systematic dataset analysis and focus exclusively on one dataset.

For metric design, the most commonly used metrics in this field are AUC (Area Under receiver operating characteristic Curve) and AP (Average Precision). These metrics treat VAD as a frame-wise binary classification task, which is insensitive to the temporal position of predictions. 
In safety-critical applications such as emergency response or fall detection systems, low anomaly detection latency \footnote{The \textit{latency} in this paper denotes how rapidly scores rise after an anomaly occurs rather than computational throughput.} is vital for triggering actionable alerts.
Figure~\ref{fig:motivation}(b) illustrates two detection curves achieving identical AUC/AP, but the first curve exhibits an earlier detection compared to the second.
Although there are metrics that address detection latency in time series anomaly detection \cite{TSAD_review,NAB_metric} and continual VAD \cite{ContinualVAD}, they assume anomalies as isolated points rather than temporal events, rendering them inapplicable to generic VAD evaluation.

For benchmarking data, current large-scale datasets primarily collect videos from the internet or movies with various scenes. As exemplified in Figure~\ref{fig:motivation}(c), a common issue arises when the test scenes are visually similar to those in the training data but contain different events. Such test data are too easy for the model and fail to reveal whether it has overfitted to the scene (e.g., treating a crowded grocery store as abnormal). The ideal test data should comprise videos of the same scene with different events (e.g., shopping in a cramped store peacefully). However, these data are not available on the web.
Though some datasets \cite{Avenue,Subway,NWPU,ubnormal,ShanghaiTech} collected by manual recording or 3D animations contain different behaviors in the same scene, they either lack severe anomalies (e.g., explosions), exhibit domain gaps from real-world data, or suffer from the limited scale of dataset samples. 
\IEEEpubidadjcol

In this paper, we focus on better evaluation methods for VAD algorithms.
\textbf{First}, we systematically analyze the three aforementioned limitations on three large-scale VAD datasets (UCF-Crime \cite{ucf-crime}, XD-Violence \cite{xdviolence}, MSAD \cite{msad}) through multi-round re-annotation (i.e., each round is labeled by different annotators) and exploratory experiments.
\textbf{Second}, we introduce probabilistic AUC/AP metrics (Prob AUC/AP) to mitigate annotation bias and propose a novel Latency-aware Average Precision (LaAP) metric to evaluate the latency of VAD algorithms. 
The probabilistic AUC/AP aggregates multiple rounds of annotations into probability labels, then extends the confusion matrix to the cumulative probabilities, and calculate AUC/AP based on that.
The LaAP replaces AP's Precision-Recall curve with a Precision-LaRecall curve, where LaRecall rewards earlier detections of anomalies through a time-decaying weighting mechanism. 
\textbf{Third}, we construct two synthetic benchmarks (UCF-HN \& MSAD-HN) via image-to-video diffusion models \cite{vidu,wan21} to evaluate overfitting in VAD algorithms. 
UCF-HN and MSAD-HN comprise normal videos generated from keyframes of abnormal scenes in UCF-Crime and MSAD training sets, where high false alarm rates (FAR) on these data reflect overfitting.

Extensive experiments validate the reliability of the proposed metrics and the realism of the proposed synthetic benchmarks. We report performance comparisons on ten state-of-the-art (SOTA) open-source VAD models \cite{RTFM,MGFN,URDMU,PEL,vadclip,TEVAD,CLIP-TSA,LAVAD,GlanceVAD,HolmesVAU} with our new evaluation methods.
The comparisons reveal two critical findings: 
1) Some methods demonstrating strong performance under conventional metrics exhibit degraded performance on our proposed metrics, suggesting overfitting to annotation bias or having large latency;
2) Existing methods have high false alarm rates on the new benchmarks, suggesting vulnerable generalization.
Our contributions are summarized as follows:

\begin{enumerate}
\item We introduce the Prob-AUC/AP metrics and propose the LaAP metric, which mitigate annotation bias and reward early anomaly detection.
\item We construct the UCF-HN and MSAD-HN benchmarks with hard normal videos generated via diffusion models to evaluate scene overfitting in VAD methods.
\item We report results for SOTA VAD methods using our proposed evaluation framework and provide new perspectives into model behaviors.
\end{enumerate}

\section{Related Works}

\subsection{Video Anomaly Detection Metrics}

The majority of existing works employ AUC and AP as evaluation metrics. Here, AUC represents the area under the TPR (True Positive Rate)-FPR (False Positive Rate) curve, while AP denotes the area under the Precision-Recall curve. Both metrics comprehensively evaluate the performance under varying decision thresholds, with AUC being insensitive to class imbalance \cite{MCC_metric} and AP demonstrating better capability for positive instances \cite{xdviolence}. 
In video anomaly detection, anomalies typically manifest as continuous multi-frame events, and early detection of such anomalies is often prioritized in practical applications. However, both AUC and AP treat predictions as independent instances, thereby neglecting the event-based nature of anomalies and exhibiting insensitivity to detection latency. While the NAB metric \cite{NAB_metric} in time-series anomaly detection and the APD metric \cite{ContinualVAD} in continual learning VAD incorporate prediction latency, they model anomalies as discrete points and measure delays within a fixed time window, which overlook variations in duration of abnormal events, rendering them inapplicable for general VAD evaluation.
Moreover, existing metrics only utilize single-round annotation, which would introduce annotator bias, as mentioned in some works \cite{ECCV_CrossDomain_LimitedSup,MultiDomainVAD,lagovad} and systematically analyzed in this paper.

To address these issues, we propose LaAP, which rewards early anomaly detection through an adaptive time window and nonlinear score functions, offering a novel perspective for VAD evaluation. We also conduct independent re-annotations for three large-scale datasets and introduce ProbAUC and ProbAP to mitigate individual annotator bias.

\subsection{Video Anomaly Detection Benchmarks}
Table~\ref{tab:dataset} compares existing VAD datasets. 
Some datasets \cite{Subway,Ped,Avenue,ShanghaiTech,NWPU} manually record videos with actors performing abnormal events, but they are limited in scale and lack severe anomaly events (e.g., explosions). 
In contrast, \cite{ubnormal} obtains videos using synthetic 3D animations, which is able to provide fine-grained pixel-level labels. Nevertheless, it introduces significant discrepancies from real-world data, limiting its applicability.

To address these limitations, datasets like \cite{ucf-crime,xdviolence,msad} collect videos from the web and movies, greatly expanding the range of anomaly categories and increasing dataset scale. 
UCF-Crime \cite{ucf-crime} contains 1,900 surveillance videos from the web, covering 13 crime-related categories. 
XD-Violence \cite{xdviolence} is another large-scale dataset with 4,754 videos with an audio channel, which spans 6 violence-related categories. Its videos primarily come from movies, news, and user-uploaded content on the web, often involving significant editing. 
MSAD \cite{msad} is a more recent dataset that covers 14 distinct scenarios and includes 11 base categories, with videos of higher resolution. The anomalies in MSAD range from human-related events, such as people falling, to non-human-related anomalies.
However, these datasets do not provide both normal and abnormal videos for the same scene, which makes it difficult to evaluate whether a model has overfitted to specific scenes rather than learning to detect true anomaly patterns. For example, a dataset might contain a video of a van exploding at a gas station but lacks a normal video showing the same van refueling and departing safely from the same location. An overfitting model might mistakenly treat the gas station or the van as anomalies. 

In this work, we construct UCF-HN and MSAD-HN, generated via image-to-video diffusion models, which synthesize normal behaviors under identical scenes to evaluate scene overfitting.

\subsection{Synthesizing Videos in Video Anomaly Detection}
Due to the sparsity of anomalies in reality, some works have explored data synthesis.
GTACrash \cite{GTACrash} acquires crash data from gaming environments, exhibiting limited physical realism.
UBNormal \cite{ubnormal} mploys 3D animation, demonstrating suboptimal human motion naturalness and environmental integration.
OVVAD \cite{OVVAD} employs a text-to-image diffusion model coupled with an image animation model for frame generation, resulting in conspicuous visual artifacts and temporal inconsistencies.
Astrid et al. \cite{Astrid2021} and Narwade et al. \cite{NarwadeICPR24} generate anomalies through cut-and-paste operations, leading to significant background inconsistencies.
Apart from the issues related to the quality of synthesizing, these methods only focus on using the synthesized data for training. 

In contrast, this work emphasizes using synthetic data to evaluate scene overfitting in scenarios where obtaining real-world data is extremely challenging. We also employ state-of-the-art video generation models and incorporate manual efforts to ensure high-quality synthesis.

\section{Analysis of Current Metrics and Benchmarks}
\label{sec:analysis}

\subsection{Bias Prevails in Annotations}
\label{sec:analysis_bias}

\begin{table}[t]
  \centering
  \caption{Comparisons of VAD datasets. \textit{Intra-Scene Pos\&Neg} denotes the inclusion of both normal and abnormal videos of the same scene.}
  \label{tab:dataset}
  \setlength{\tabcolsep}{1.5mm}
    \begin{tabular}{@{}llcc@{}}
    \toprule
    Dataset & Source & \# videos & Intra-Scene Pos\&Neg \\ \midrule
    Subway \cite{Subway}       & recorded   & 2    & \checkmark \\
    UCSD Ped \cite{Ped}        & recorded   & 98   & \checkmark \\
    Avenue \cite{Avenue}       & recorded   & 35   & \checkmark \\
    ShanghaiTech \cite{ShanghaiTech} & recorded   & 437  & \checkmark \\
    NWPU Campus \cite{NWPU}    & recorded   & 547  & \checkmark \\
    UBNormal \cite{ubnormal}   & synthesized & 543  & \checkmark \\ \midrule
    UCF-Crime \cite{ucf-crime} & web         & 1900 & \ding{55}  \\
    XD-Violence \cite{xdviolence} & web, movie & 4754 & \ding{55}  \\
    MSAD \cite{msad}           & web         & 720  & \ding{55}  \\ \bottomrule
    \end{tabular}%
\end{table}

The validity of evaluation metrics depends on the precision of annotations. In this section, we systematically analyze annotation discrepancies by re-annotating three large-scale datasets: UCF-Crime, XD-Violence, and MSAD.
We conduct frame-level re-annotation of temporal boundaries for all abnormal videos in the test sets, requiring annotators to strictly follow the official definitions of each dataset. For example, in UCF-Crime, the category ``Burglary'' is defined as: \textit{This event contains videos that show people (thieves) entering into a building or house with the intention to commit theft. It does not include use of force against people.} 
Each annotation round is assigned to a different annotator for independent labeling.
In total, we obtain four independent rounds of annotations per dataset, including the original ones. Notably, for UCF-Crime, one annotation round is adopted from \cite{ECCV_CrossDomain_LimitedSup}.

\begin{figure}[tbp]
    \centering
    \includegraphics[width=\linewidth]{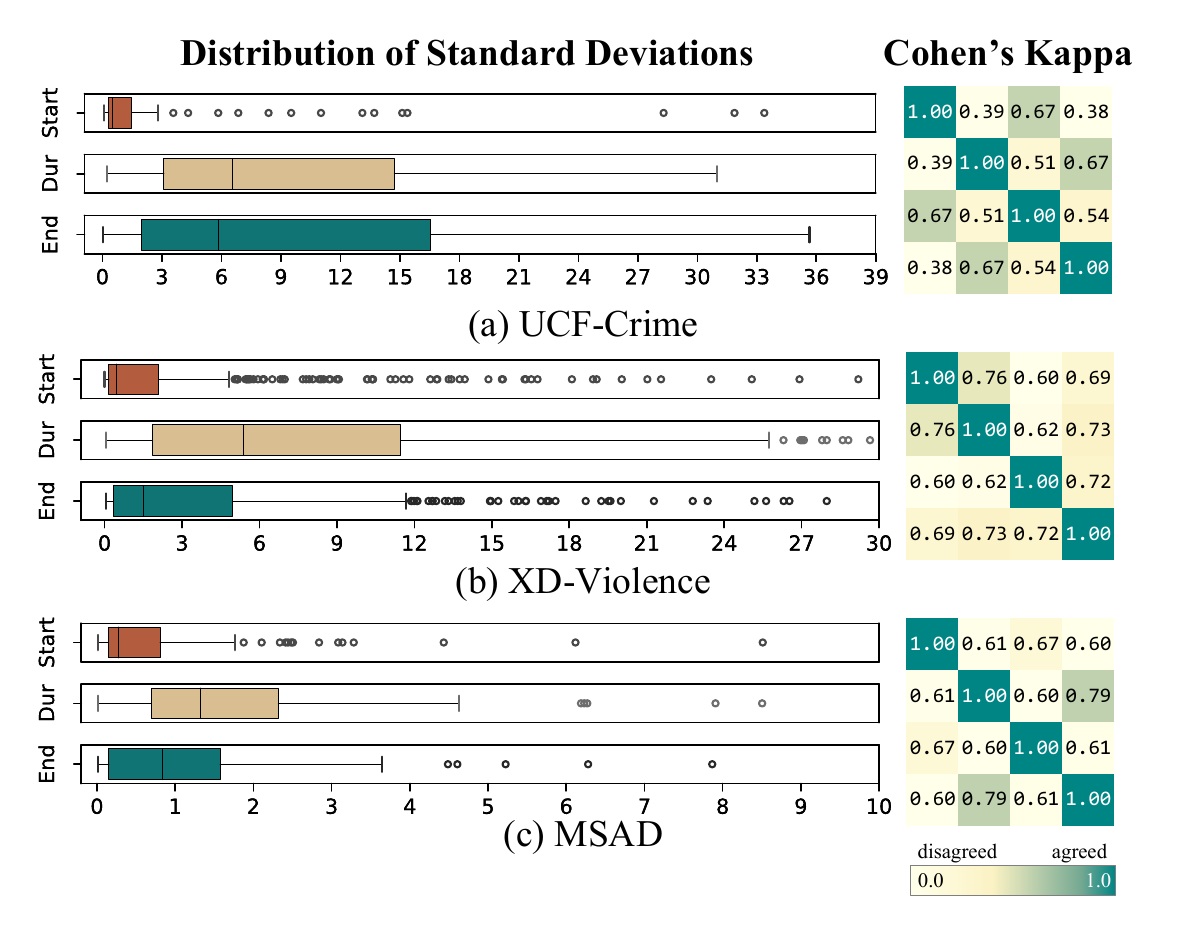}
    \caption{\textbf{Left}: The distribution of standard deviations for start time, duration, and end time (in seconds) across different annotations in three datasets. \textbf{Right}: The pair-wise Cohen's Kappa between annotators.}
    \label{fig:dataset-disagreement}
\end{figure}
\begin{table}[tbp]
  \centering
  \caption{The overall Fleiss' Kappa and the median of standard deviations for start time, duration, and end time (in seconds).}
  \label{tab:dataset-deviation}
    \begin{tabular}{@{}lccc@{}}
    \toprule
     & UCF & XD & MSAD \\ \midrule
    Fleiss' Kappa $\uparrow$ & 0.51 & 0.68 & 0.64 \\ 
    median of start std $\downarrow$ & 0.50 & 0.44 & 0.27 \\
    median of duration std $\downarrow$ & 6.56 & 5.39 & 1.32 \\
    median of end std $\downarrow$ & 5.84 & 1.50 & 0.83 \\ \bottomrule
    \end{tabular}%
\end{table}

During this process, we observe that some abnormal videos in UCF-Crime exhibit category ambiguity. To address this, we further perform categorical re-annotation of all abnormal videos in UCF-Crime. Importantly, annotators were not informed that only abnormal videos were selected, and the annotators responsible for categorical re-annotation were different from those performing temporal boundary labeling, thereby mitigating potential bias.

We quantify the overall frame-level annotation consistency using Cohen's Kappa \cite{cohen1960kappa} and Fleiss' Kappa \cite{fleiss1971kappa} in Figure~\ref{fig:dataset-disagreement} and Table~\ref{tab:dataset-deviation}. Both metrics range in $[-1, 1]$ where values approaching 1 indicate perfect agreement. 
The results reveal that the lowest inter-annotator Cohen's Kappa is 0.382, with overall Fleiss' Kappa ranging from 0.51 to 0.68, indicating substantial annotation discrepancies.
To identify the source of discrepancies, we analyze the distribution of standard deviations of start time, duration, and end time in Figure~\ref{fig:dataset-disagreement} and Table~\ref{tab:dataset-deviation}.
Note that the start and end times correspond to the earliest and latest annotated abnormal frames in a video, and the duration refers to the total time of all annotated abnormal segments in a video.

\noindent\textbf{Finding 1.} 
The results reveal that the deviations for the duration and end time are significantly higher than those for the start time, which indicates that annotators diverge in granularity (i.e., whether to label the most salient segments or more comprehensive events) and determination of ending (i.e., whether to label the aftermath).

\begin{figure}[tbp]
    \centering
    \includegraphics[width=\linewidth]{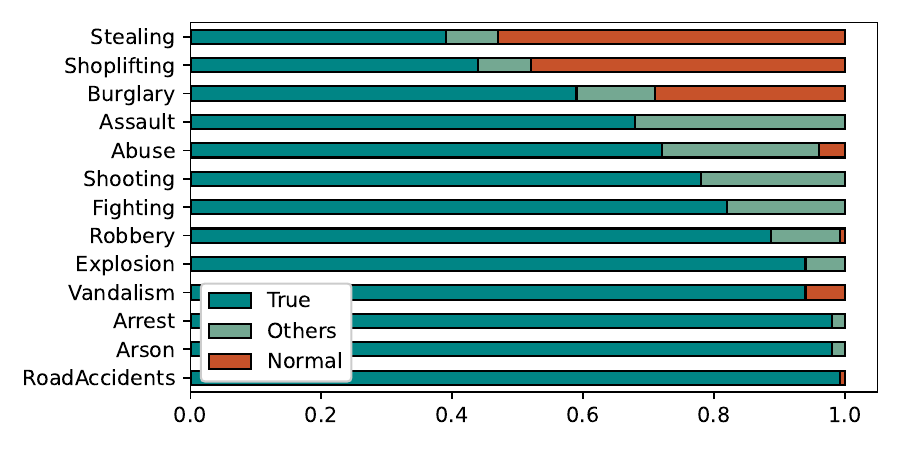}
    \caption{Visualization of discrepancies in UCF-Crime \cite{ucf-crime}. \colorbox{MyDarkGreen}{True} indicates agreement between annotations; \colorbox{MyLightGreen}{Others} denotes our annotation assigning the sample to another anomaly category; \colorbox{MyDarkRed}{Normal} denotes videos classified as normal.}
    \label{fig:cls_ambiguity}
\end{figure}

For categorical ambiguity, we visualizes the discrepancies between the original annotation and ours in UCF-Crime in Fig.~\ref{fig:cls_ambiguity}.
Videos originally labeled as \textit{Stealing}, \textit{Shoplifting}, and \textit{Burglary} exhibit high rates of ``misclassification'' as \textit{Normal} ($\geq 30\%$). We attribute this to inherent dependence on contextual information beyond the video. 
For instance, it is challenging to distinguish between the staff member’s routine actions and a shoplifter’s behavior of taking away a t-shirt in surveillance footage without knowing who the staff member is. Such ambiguous instances would introduce confounding signals during modeling and evaluating.
Visualization examples are given in the supplementary.

\noindent\textbf{Finding 2.} 
Categories whose recognition inherently relies on contextual knowledge beyond visual evidence are less suitable for evaluation for vision-based methods.

Therefore, in the following experiments on UCF-Crime, we exclude these three ambiguous categories.

\subsection{Existing Methods Differ in Latency}
\begin{figure}[tbp]
    \centering
    \includegraphics[width=0.73\linewidth]{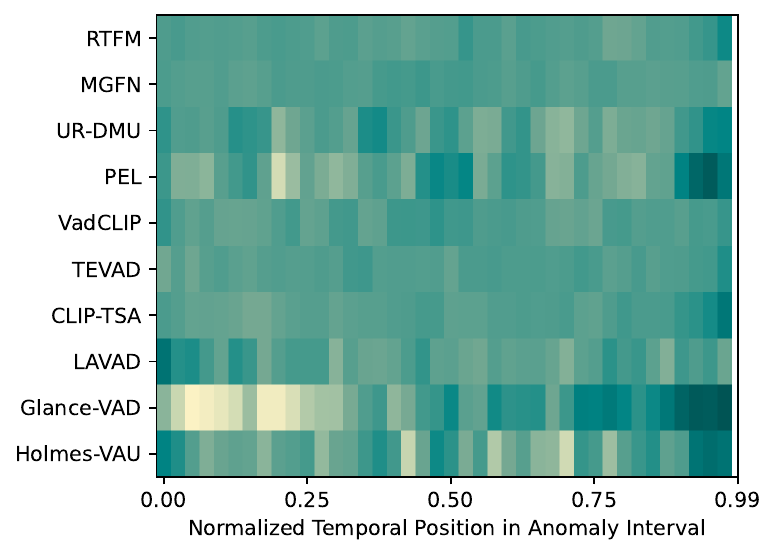}
    \caption{Heatmap of predicted anomaly positions of different models in abnormal intervals. Brighter areas indicate higher density. The introducion of these models are in Sec.~\ref{sec:baselines}.}
    \label{fig:pos_heatmap}
\end{figure}

In practical applications, accurate and early anomaly detection is preferable under equivalent model capability.
However, current metrics treat VAD task as a frame-independent binary classification task while neglecting positional information in predictions. 
Figure~\ref{fig:pos_heatmap} illustrates the distribution of prediction latency across multiple models.
We select prediction scores within abnormal intervals, binarize them, obtain the positions of positive instances, and normalize them using the length of the abnormal interval.
The results reveal that certain models (e.g., RTFM \cite{RTFM}, CLIP-TSA \cite{CLIP-TSA}) exhibit relatively uniform predictions, while some (PEL \cite{PEL}, Glance-VAD \cite{GlanceVAD}) achieve higher positive concentrations in earlier stages. 

\noindent\textbf{Finding 3.} Existing models exhibit different behaviors in prediction latency.

\subsection{Current Benchmarks Struggle to Evaluate Scene Overfitting}
\label{sec:analysis_easytest}
Existing VAD models trained on large-scale web-sourced datasets \cite{ucf-crime,xdviolence,msad} with either weak or full supervision are prone to overfitting to training scenes. However, current benchmarks are insufficient to evaluate such overfitting, as their test sets mainly consist of scenes that differ substantially from those in the training data. This makes them relatively easy for the models, while leaving open the question of whether performance would remain robust on test scenes that closely resemble the training ones.
For example, a production scenario may involve a crowded grocery store with a checkout counter on the right, which is included in the training data, whereas the existing test data only contains videos from mobile phone stores and clothing retailers.

\begin{table}[ht]
\centering
\caption{Re-splitting strategy and performance on the re-split ShanghaiTech.}
\label{tab:ShT-exp}
\resizebox{\linewidth}{!}{%
\begin{tabular}{@{}l|cc|cccc@{}}
\toprule
\multirow{3}{*}{Split} & \multicolumn{2}{c|}{Re-split Scene Number}                                                   & \multicolumn{4}{c}{Performance (\%)}                                   \\ \cmidrule(l){2-7} 
                       & \multicolumn{1}{l}{\multirow{2}{*}{Normal}} & \multicolumn{1}{l|}{\multirow{2}{*}{Abnormal}} & \multicolumn{2}{c|}{PEL \cite{PEL}}                 & \multicolumn{2}{c}{VadCLIP \cite{vadclip}} \\
                       & \multicolumn{1}{l}{}                        & \multicolumn{1}{l|}{}                          & AUC $\uparrow$   & \multicolumn{1}{c|}{FAR$_{0.5}  \downarrow$} & AUC $\uparrow$      & FAR$_{0.5} \downarrow$      \\ \midrule
train set              & 01--06                                      & 07--10                                         & --    & --                               & --        & --              \\
easy test set          & 11,12,13                                    & 11,12,13                                       & 94.55 & 0.38                             & 77.13     & 17.53           \\
hard test set          & 07--10                                      & 01--06                                         & 16.35 & 44.10                            & 36.46     & 53.86           \\ \bottomrule
\end{tabular}%
}
\end{table}
Therefore, we conduct a preliminary experiment by re-splitting the ShanghaiTech dataset \cite{ShanghaiTech} as shown in Table~\ref{tab:ShT-exp}. ShanghaiTech is a manually recorded dataset with normal and abnormal videos across 13 scenes. 
We first construct a training set and a easy test set to simulate the case of web-sourced datasets, where each scene in the training set contains exclusively normal or exclusively abnormal videos, with test scenes being distinct from those in the training set. Subsequently, we construct a hard test set containing identical scenes to the training set but with reversed video normality in the same scenes.
We train two SOTA models \cite{PEL, vadclip} on the training set, evaluate on both test sets with the best checkpoint. 
As shown in Table~\ref{tab:ShT-exp}, while the models demonstrate satisfactory performance on the easy test set, which simulates existing benchmarks, they significantly deteriorate on the hard test set, resulting in a sharp AUC dropping (16\%/36\%) with a high FAR (44\%/54\%), indicating severe scene overfitting.

\noindent\textbf{Finding 4.} Although existing weakly/fully supervised methods achieve strong results on easy test sets with disjoint scenes, they have the potential risk of scene overfitting.

\section{Proposed Metrics and Benchmarks}

\subsection{Preliminaries}

We introduce the calculations of the traditional VAD metrics AUC and AP as follows.
First, the predicted probabilities $\hat{y}_i^j$ and binary ground truths $y_i^j$ of all videos are concatenated into two long sequences:
\begin{align}
    \mathbf{\hat{y}} &= [\hat{y}_1^1, \hat{y}_2^1, \dots, \hat{y}_{M_1}^1, \hat{y}_1^2, \dots, \hat{y}_{M_V}^V] \in {[0,1]}^{N}, \\
    \mathbf{y} &= [y_1^1, y_2^1, \dots, y_{M_1}^1, y_1^2, \dots, y_{M_V}^V] \in {\{0,1\}}^{N},
\end{align}
where $j$ indexes videos and $i$ indexes frames within each video. $M_j$ is the frame count of video $j$, and $V$ is the total number of videos.
Then, given a threshold $\tau$, the items of the confusion matrix are defined as:
\begin{align}
\mathrm{TP} &= \textstyle\sum_{j=1}^V \sum_{i=1}^{M_j} \mathds{1}(\hat{y}_i^j \ge \tau)\, y_i^j, \label{eq:confusion_mat_start} \\
\mathrm{FP} &= \textstyle\sum_{j=1}^V \sum_{i=1}^{M_j} \mathds{1}(\hat{y}_i^j \ge \tau)\, (1-y_i^j), \\
\mathrm{FN} &= \textstyle\sum_{j=1}^V \sum_{i=1}^{M_j} \mathds{1}(\hat{y}_i^j < \tau)\, y_i^j, \\
\mathrm{TN} &= \textstyle\sum_{j=1}^V \sum_{i=1}^{M_j} \mathds{1}(\hat{y}_i^j < \tau)\, (1-y_i^j). \label{eq:confusion_mat_end}
\end{align}
We define a general area-under-curve operator $\Psi(X,Y)$ as:
\begin{equation}
    \Psi(X,Y) = \textstyle\int^1_0 Y(\tau)\ dX(\tau),
\end{equation}
where $X,Y$ are two functions of $\tau$, called the integrator and the integrand, respectively.
AUC/AP are defined as:
\begin{align}
    \mathrm{R}=\mathrm{TPR}&=\mathrm{TP}/(\mathrm{TP}+\mathrm{FN}), \label{eq:metric_start} \\
    \mathrm{P}&=\mathrm{TP}/(\mathrm{TP}+\mathrm{FP}), \\
    \mathrm{FPR}&=\mathrm{FP}/(\mathrm{FP}+\mathrm{TN}), \label{eq:metric_end} \\
    \mathrm{AUC} = \Psi(\mathrm{FPR}, &\mathrm{TPR}), \quad \mathrm{AP} = \Psi(\mathrm{R}, \mathrm{P}).
    \label{eq:tradition-auc-ap}
\end{align}

\subsection{Probabilistic AUC and AP Metrics}
To mitigate the impact of bias, we ask: \textit{can standardizing the annotation guideline solve the problem?} We argue it cannot, primarily for two reasons: 1) the intrinsic complexity and diversity of anomalies make it costly to train annotators, especially given a large number of categories; 2) more critically, any fixed guideline induces a specific dataset bias. Consequently, an algorithm's poor performance may stem from a misalignment with this bias rather than a genuine lack of capability. Therefore, the concept of a deterministic ground truth is inadequate. Instead, a more robust approach is to define the ground truth as a distribution and approximate it by utilizing multiple annotations.

We explored three potential approaches of utilizing multiple annotations.
One straightforward approach is to aggregate annotations via a voting mechanism. Nevertheless, using hard labels incurs a loss of information and cannot avoid bias.
We also tried calculate AUC/AP separately across annotations and use the average as a more reliable metric. However, a significant limitation is the lack of a clear upper bound, as a prediction would need to perfectly match all annotations to achieve a score of 1, which is impossible because the annotations are different from each other.
Consequently, inspired by \cite{ARatio}, we adopted the approach of first averaging multi-round annotations to create a soft label, and then extend the traditional AUC/AP to probabilistic AUC/AP (i.e., ProbAUC, ProbAP) to utilize the soft labels, which avoids the aforementioned drawbacks.

\begin{figure}[tbp]
    \centering
    \includegraphics[width=0.9\linewidth]{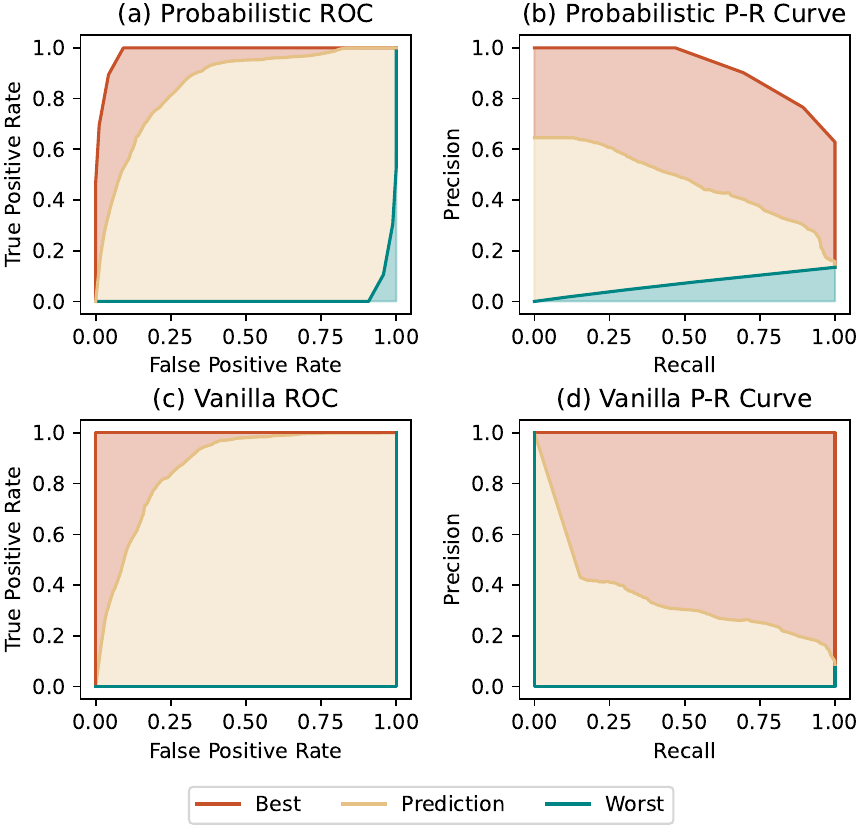}
    \caption{Comparison between vanilla ROC/P-R curve and probabilistic ROC/P-R curve.}
    \label{fig:prob_curve_vis}
    \vspace{-1em}
\end{figure}

Formally, ProbAUC and ProbAP extend the conventional evaluation framework by substituting the binary ground truth vector $\mathbf{y}$ with a probabilistic ground truth vector $\tilde{\mathbf{y}}$:
\begin{equation}
    \tilde{\mathbf{y}} = [\tilde{y}_1^1, \tilde{y}_2^1, \dots, \tilde{y}_{M_1}^1, \tilde{y}_1^2, \dots, \tilde{y}_{M_V}^V] \in {\{0,1\}}^{N},
\end{equation}
Subsequently the components of the confusion matrix and the related base metrics can be derived in a manner analogous to Eqs.~\eqref{eq:confusion_mat_start}-\eqref{eq:confusion_mat_end},\eqref{eq:metric_start}-\eqref{eq:metric_end}.
However, the introduction of probabilistic labels alters the effective range of the performance metric. As illustrated in Fig.~\ref{fig:prob_curve_vis}, the area under the curve corresponding to the best classifier is no longer one, while that of the worst classifier deviates from zero. Accordingly, we introduce ProbAUC and ProbAP with a normalized definition:

\begin{align}
    \mathrm{ProbAUC} &= \frac{\Psi(\mathrm{FPR},\mathrm{TPR})-\Psi(\mathrm{FPR^{\textcolor{MyDarkGreen}{w}}},\mathrm{TPR^{\textcolor{MyDarkGreen}{w}}})}{\Psi(\mathrm{FPR^{\textcolor{MyDarkRed}{b}}},\mathrm{TPR^{\textcolor{MyDarkRed}{b}}})-\Psi(\mathrm{FPR^{\textcolor{MyDarkGreen}{w}}},\mathrm{TPR^{\textcolor{MyDarkGreen}{w}}})}, \\
    \mathrm{ProbAP} &= 
    \frac
    {\Psi(\mathrm{R}, \mathrm{P})-\Psi(\mathrm{R^{\textcolor{MyDarkGreen}{w}}}, \mathrm{P^{\textcolor{MyDarkGreen}{w}}})}
    {\Psi(\mathrm{R^{\textcolor{MyDarkRed}{b}}}, \mathrm{P^{\textcolor{MyDarkRed}{b}}})-\Psi(\mathrm{R^{\textcolor{MyDarkGreen}{w}}}, \mathrm{P^{\textcolor{MyDarkGreen}{w}}})},
    \label{eq:prob-auc-ap}
\end{align}
where the superscript $w$ denotes the worst classifier and $b$ denotes the best classifier.
From a geometric perspective, the metric can be interpreted as:
\begin{equation}
    \mathrm{ProbAUC/AP} = \frac{\colorbox{MyDarkYellow}{\text{YellowArea}}}{\colorbox{MyDarkYellow}{\text{YellowArea}}+\colorbox{MyDarkRed}{\text{RedArea}}}.
\end{equation}
Notably, when $\tilde{\mathbf{y}}$ is reduced to a hard label, ProbAUC and ProbAP degenerate to the conventional AUC and AP, i.e., $\Psi(\mathrm{FPR^{\textcolor{MyDarkGreen}{w}}},\mathrm{TPR^{\textcolor{MyDarkGreen}{w}}})=\Psi(\mathrm{R^{\textcolor{MyDarkGreen}{w}}}, \mathrm{P^{\textcolor{MyDarkGreen}{w}}})=0$ and $\Psi(\mathrm{FPR^{\textcolor{MyDarkRed}{b}}},\mathrm{TPR^{\textcolor{MyDarkRed}{b}}})=\Psi(\mathrm{R^{\textcolor{MyDarkRed}{b}}}, \mathrm{P^{\textcolor{MyDarkRed}{b}}})=1$.

\subsection{Latency-aware AP Metric}
\begin{figure}[tbp]
    \centering
    \includegraphics[width=0.9\linewidth]{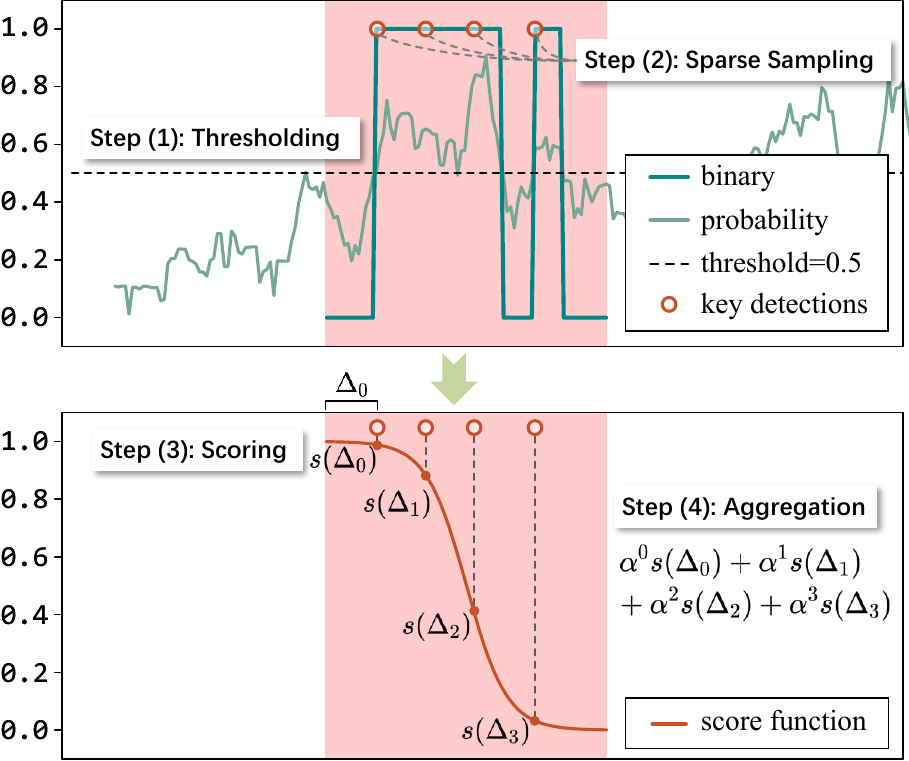}
    \caption{Visualization of the computational steps of LaRecall.}
    \label{fig:LaRecall}
\end{figure}

Motivated by the latency sensitivity of real-world anomaly detection systems, we introduce Latency-aware Average Precision (LaAP), which expands the standard AP metric through a time-decaying weighting mechanism, explicitly rewarding earlier predictions.
Formally, LaAP is defined as the area under the Precision-LaRecall curve under all possible thresholds $\tau$:
\begin{equation}
\mathrm{LaAP} = \int_0^1 \text{LaRecall}(\tau) \ d\mathrm{P}(\tau),
\end{equation}
where the LaRecall replaces Recall in AP and is the core component.
As illustrated in Figure~\ref{fig:LaRecall}, the computation of LaRecall involves four steps:

\noindent (1) \textbf{Thresholding}: Convert predicted probabilities to binary labels given a threshold:
\begin{equation}
    \overline{Y}^j = \{ \overline{y}^j_i\}_{i=1}^{M_j}, \quad
    \overline{y}_i^j = \mathds{1}(\hat{y}_i^j \geq \tau),
\end{equation}
where $\overline{Y}^j$ is the binary predictions of the j-th video.

\noindent (2) \textbf{Sparse Sampling}: Sample multiple key detection frames from positive predictions, ensuring temporal spacing $\phi$ between consecutive detections:
\begin{equation}
    \mathcal{A}^j = \text{Sample}(\overline{Y}^j), \quad \mathcal{A}^j_{k+1}-\mathcal{A}^j_{k} > \phi,
\end{equation}
where $\mathcal{A}$ denotes the indices of sampled frames and the subscript denotes $k$-th sampled frame.
The selection of multiple detections helps to reduce the influence of noise, and the spacing is set to avoid repeated scoring.

\noindent (3) \textbf{Scoring}: Apply a Sigmoid-like time-decaying scoring function to the sampled detections, which can give a score close to 1 during a short period after the anomaly occurs, followed by a smooth decrease to 0. This property aligns with the intuition of rewarding early detection. Formally,
\begin{equation}
    s(\Delta^j_k) = 1 - \frac{1}{1+\exp(-\beta(2\Delta^j_k-1))}, \quad
    \Delta^j_k = \frac{\mathcal{A}^j_{k}-t_{\text{start}}}{t_{\text{end}}-t_{\text{start}}},
\end{equation}
where $t_{\text{start}}, t_{\text{end}}$ denote the indices of the start frame and the end frame, respectively, and $\beta$ controls the curve shape.

\noindent (4) \textbf{Aggregation}: Compute video-level LaRecall through exponentially decaying weighted function $w(k) = \alpha^{k}$, and then average across all abnormal videos. Specifically, if a video yields no positives under a given detection threshold, the corresponding LaRecall score is set to zero.
\begin{equation}
    \text{LaRecall} = \sum_{j=0}^{N_a} 
    \frac
    {\sum_{k=0}^{|\mathcal{A}^j|-1}{w(k) s(\Delta^j_k)}}
    {\sum_{k=0}^{|\mathcal{A}^j|-1} w(k)},
\end{equation}
where $N_a$ is the number of abnormal events in the dataset, $|\mathcal{A}^j|$ denotes the number of sampled frames, $\Delta^j_k$ denotes the normed position of the $k$-th detection in the $j$-th video , and $\alpha$ are parameters controlling decaying degree.

In practice, LaAP assumes that only one abnormal event exists in a video due to the sparsity of anomalies.
In existing datasets, almost all data satisfy this assumption except for the XD-Violence dataset, which has many heavily edited videos and contains multiple logically unrelated events within a single video. Consequently, we do not compute the LaAP metric on the XD-Violence dataset. For future datasets containing multiple independent anomalies in a single video, LaAP can be applied by splitting the video into segments.
For multi-round annotations, we select the median of the first abnormal frames and the latest abnormal frames across different annotations as the starting and ending times. This also mitigates the single annotation bias issue discussed in Section~\ref{sec:analysis_bias}.

We provide guidelines and default settings for three parameters as follows:
\begin{itemize}
    \item $\phi$: This parameter is to avoid redundant scoring, which is a positive integer representing the sampling interval. It should remain fixed at 16 for most cases.

    \item $\alpha$: This parameter controls the relative importance of sampling points, which should be a positive number greater than 1. A higher $\alpha$ reduces attention to later sampling points. With our default $\alpha = 2$, the weights for points beyond six points decay below 1\%. To prioritize earlier detections, $\alpha$ should be increased.

    \item $\beta$: This parameter controls the steepness of the Sigmoid-like scoring curve, determining LaAP's tolerance near event boundaries, which takes a positive number. As $\beta$ approaches 0, the curve exhibits more linearity. And larger values make it approximate a step function. With our default $\beta = 7$, sampling points within the 34\% interval preceding anomaly onset receive scores above 0.9. To allow for greater detection latency, $\beta$ can be increased.
\end{itemize}

\subsection{Video-Diffusion-Generated Hard Normal Benchmarks}
Having observed the severe overfitting in the re-split ShanghaiTech dataset, we raise the question: \textit{Does the scene overfitting persist when training data contains more scenes?}
However, validation becomes extremely challenging due to the inability to acquire the same-scene data in the real world. Therefore, leveraging advancements in AIGC technologies, we synthesize such data using video diffusion models \cite{vidu,wan21}, which form two new hard normal benchmarks: UCF-HN and MSAD-HN.

\begin{table*}[t]
\centering
  \caption{Comparisons between various state-of-the-art VAD models with original AUC/AP, our proposed ProbAUC/ProbAP, and our proposed LaAP metric. Note that the metrics on UCF-Crime exclude three categories as discussed in Section~\ref{sec:analysis_bias}. The scores are colored according to the rank. $\ast$ denotes zero-shot methods. $\dagger$ denotes methods with extra supervision signals.}
  \label{tab:benchmarking-1}
\newcolumntype{C}{>{\centering\arraybackslash}X} %

\resizebox{\textwidth}{!}{%
\begin{tabularx}{\textwidth}{@{}l|*{5}{C}|*{4}{C}|*{5}{C}@{}}
\toprule
\multicolumn{1}{c|}{}                        & \multicolumn{5}{c|}{UCF-Crime}                                                                                                                                & \multicolumn{4}{c|}{XD-Violence}                                                                                              & \multicolumn{5}{c}{MSAD}                                                                                                                                      \\ 
\multicolumn{1}{c|}{\multirow{-2}{*}{Model}} & AUC                           & ProbAUC                       & AP                            & ProbAP                        & LaAP                          & AUC                           & ProbAUC                       & AP                            & ProbAP                        & AUC                           & ProbAUC                       & AP                            & ProbAP                        & LaAP                          \\ \midrule
Random                                       & \cellcolor[HTML]{D9D9D9}0.499 & \cellcolor[HTML]{D9D9D9}0.500 & \cellcolor[HTML]{D9D9D9}0.075 & \cellcolor[HTML]{D9D9D9}0.135 & \cellcolor[HTML]{D9D9D9}0.129 & \cellcolor[HTML]{D9D9D9}0.500 & \cellcolor[HTML]{D9D9D9}0.500 & \cellcolor[HTML]{D9D9D9}0.231 & \cellcolor[HTML]{D9D9D9}0.259 & \cellcolor[HTML]{D9D9D9}0.497 & \cellcolor[HTML]{D9D9D9}0.498 & \cellcolor[HTML]{D9D9D9}0.228 & \cellcolor[HTML]{D9D9D9}0.213 & \cellcolor[HTML]{D9D9D9}0.267 \\
RTFM \cite{RTFM}                             & \cellcolor[HTML]{C9DCB7}0.835 & \cellcolor[HTML]{B0D2B1}0.864 & \cellcolor[HTML]{C9DCB7}0.239 & \cellcolor[HTML]{C9DCB7}0.434 & \cellcolor[HTML]{E2E7BE}0.480 & -                             & -                             & -                             & -                             & \cellcolor[HTML]{97C6AB}0.868 & \cellcolor[HTML]{64B19E}0.870 & \cellcolor[HTML]{64B19E}0.662 & \cellcolor[HTML]{64B19E}0.665 & \cellcolor[HTML]{64B19E}0.713 \\
MGFN \cite{MGFN}                             & \cellcolor[HTML]{E2E7BE}0.820 & \cellcolor[HTML]{97C6AB}0.868 & \cellcolor[HTML]{E2E7BE}0.216 & \cellcolor[HTML]{E2E7BE}0.428 & \cellcolor[HTML]{329B92}0.761 & -                             & -                             & -                             & -                             & \cellcolor[HTML]{C9DCB7}0.850 & \cellcolor[HTML]{C9DCB7}0.849 & \cellcolor[HTML]{97C6AB}0.633 & \cellcolor[HTML]{97C6AB}0.626 & \cellcolor[HTML]{C9DCB7}0.664 \\
PEL \cite{PEL}                               & \cellcolor[HTML]{4BA598}0.890 & \cellcolor[HTML]{7DBBA4}0.875 & \cellcolor[HTML]{4BA598}0.305 & \cellcolor[HTML]{64B19E}0.495 & \cellcolor[HTML]{64B19E}0.727 & \cellcolor[HTML]{23948E}0.949 & \cellcolor[HTML]{6BB4A0}0.944 & \cellcolor[HTML]{23948E}0.855 & \cellcolor[HTML]{6BB4A0}0.856 & \cellcolor[HTML]{64B19E}0.873 & \cellcolor[HTML]{008585}0.873 & \cellcolor[HTML]{329B92}0.676 & \cellcolor[HTML]{008585}0.683 & \cellcolor[HTML]{329B92}0.729 \\
UR-DMU \cite{URDMU}                          & \cellcolor[HTML]{7DBBA4}0.887 & \cellcolor[HTML]{64B19E}0.889 & \cellcolor[HTML]{64B19E}0.275 & \cellcolor[HTML]{97C6AB}0.489 & \cellcolor[HTML]{97C6AB}0.693 & \cellcolor[HTML]{6BB4A0}0.940 & \cellcolor[HTML]{48A497}0.950 & \cellcolor[HTML]{6BB4A0}0.817 & \cellcolor[HTML]{48A497}0.881 & \cellcolor[HTML]{008585}0.878 & \cellcolor[HTML]{008585}0.873 & \cellcolor[HTML]{008585}0.697 & \cellcolor[HTML]{329B92}0.678 & \cellcolor[HTML]{008585}0.729 \\
VadCLIP \cite{vadclip}                       & \cellcolor[HTML]{64B19E}0.889 & \cellcolor[HTML]{329B92}0.908 & \cellcolor[HTML]{B0D2B1}0.254 & \cellcolor[HTML]{329B92}0.542 & \cellcolor[HTML]{4BA598}0.740 & \cellcolor[HTML]{48A497}0.947 & \cellcolor[HTML]{23948E}0.956 & \cellcolor[HTML]{48A497}0.845 & \cellcolor[HTML]{23948E}0.885 & \cellcolor[HTML]{329B92}0.878 & \cellcolor[HTML]{97C6AB}0.869 & \cellcolor[HTML]{C9DCB7}0.597 & \cellcolor[HTML]{C9DCB7}0.557 & \cellcolor[HTML]{97C6AB}0.665 \\
TEVAD \cite{TEVAD}                           & \cellcolor[HTML]{97C6AB}0.874 & \cellcolor[HTML]{C9DCB7}0.861 & \cellcolor[HTML]{97C6AB}0.265 & \cellcolor[HTML]{B0D2B1}0.455 & \cellcolor[HTML]{B0D2B1}0.599 & \cellcolor[HTML]{B3D3B2}0.923 & \cellcolor[HTML]{B3D3B2}0.917 & \cellcolor[HTML]{90C3A9}0.798 & \cellcolor[HTML]{90C3A9}0.829 & -                             & -                             & -                             & -                             & -                             \\
CLIP-TSA \cite{CLIP-TSA}                     & \cellcolor[HTML]{19908B}0.900 & \cellcolor[HTML]{19908B}0.920 & \cellcolor[HTML]{7DBBA4}0.272 & \cellcolor[HTML]{4BA598}0.526 & \cellcolor[HTML]{7DBBA4}0.695 & \cellcolor[HTML]{90C3A9}0.932 & \cellcolor[HTML]{90C3A9}0.936 & \cellcolor[HTML]{B3D3B2}0.772 & \cellcolor[HTML]{B3D3B2}0.820 & -                             & -                             & -                             & -                             & -                             \\
LAVAD$^\ast$ \cite{LAVAD}                    & \cellcolor[HTML]{B0D2B1}0.856 & \cellcolor[HTML]{E2E7BE}0.847 & \cellcolor[HTML]{19908B}0.336 & \cellcolor[HTML]{7DBBA4}0.491 & \cellcolor[HTML]{C9DCB7}0.592 & \cellcolor[HTML]{D8E3BB}0.854 & \cellcolor[HTML]{D8E3BB}0.865 & \cellcolor[HTML]{D8E3BB}0.620 & \cellcolor[HTML]{D8E3BB}0.700 & -                             & -                             & -                             & -                             & -                             \\
GlanceVAD$^\dagger$ \cite{GlanceVAD}         & \cellcolor[HTML]{008585}0.937 & \cellcolor[HTML]{4BA598}0.902 & \cellcolor[HTML]{008585}0.484 & \cellcolor[HTML]{008585}0.612 & \cellcolor[HTML]{008585}0.818 & \cellcolor[HTML]{008585}0.959 & \cellcolor[HTML]{008585}0.960 & \cellcolor[HTML]{008585}0.892 & \cellcolor[HTML]{008585}0.911 & -                             & -                             & -                             & -                             & -                             \\
HolmesVAU$^\dagger$ \cite{HolmesVAU}         & \cellcolor[HTML]{329B92}0.893 & \cellcolor[HTML]{008585}0.921 & \cellcolor[HTML]{329B92}0.321 & \cellcolor[HTML]{19908B}0.610 & \cellcolor[HTML]{19908B}0.800 & -                             & -                             & -                             & -                             & -                             & -                             & -                             & -                             & -                             \\ \bottomrule
\end{tabularx}%
}
\end{table*}

\noindent (1) \textbf{Select Keyframes}: 
To obtain high-quality images, we manually extracted key frames from referenced abnormal videos in the training sets of UCF-Crime and MSAD. Due to the complexity of abnormal actions, we only generate normal videos.
The extracted keyframes meet the following criteria: 1) No inclusion of abnormal events; 2) Absence of motion blur; 3) Minimal visible watermarks; 4) One keyframe per video. 

\noindent (2) \textbf{Generate Videos}:
We used image-to-video diffusion models (\texttt{Vidu 2.0}\cite{vidu} and \texttt{wan2.1\_i2v\_480p\_14B}\cite{wan21}) with instance-specific manual prompts to generate \textbf{normal} videos preserving identical scenes.
Each video was created by first generating a clip from the selected key frame, then producing an extended clip using the last frame of the first clip, followed by temporal concatenation. The generated videos last from 10 to 16 seconds. 
Manual prompt engineering ensured diverse event representations in generated videos. 

\noindent (3) \textbf{Recheck}:
During this phase, we re-generated videos exhibiting suboptimal quality while discarding frames with persistent synthesis artifacts. All retained videos were subjected to manual quality inspection to verify visual fidelity. This process yielded 100 hard normal samples for UCF-HN and 67 hard normal samples for MSAD-HN.
These two benchmarks are intended to complement the test sets of UCF-Crime and MSAD rather than replacing them entirely. Compared to the numbers of normal test videos of these datasets (150 in UCF-Crime, 120 in MSAD), we consider our benchmarks’ scale sufficient for their purpose. 

We provide both quantitative and qualitative validation of the realism of generated videos in the following experiments.
To perform benchmarking, the model should be trained on UCF-Crime or MSAD, and then be tested on the corresponding hard normal set. Since all test videos are normal, the false alarm rate (FAR) reflects the degree of overfitting. 

\begin{table*}[t]
\centering
  \caption{Performance comparison of SOTA VAD methods on UCF-Crime/MSAD and our proposed hard-normal variants (UCF-HN/MSAD-HN) using FAR$_\tau$ metric, where $\tau$ is the threshold. The scores are colored according to the value.}
  \label{tab:benchmarking-2}
    \begin{tabular}{@{}lc|cccc|cccc@{}}
    \toprule
    \multicolumn{1}{c}{} &  & \multicolumn{2}{c}{Original UCF Test} & \multicolumn{2}{c|}{UCF-HN} & \multicolumn{2}{c}{Original MSAD Test} & \multicolumn{2}{c}{MSAD-HN} \\ \cmidrule(l){3-10} 
    \multicolumn{1}{c}{\multirow{-2}{*}{Model}} & \multirow{-2}{*}{Supervision} & FAR$_{0.5}$ & FAR$_{0.8}$ & FAR$_{0.5}$ & FAR$_{0.8}$ & FAR$_{0.5}$ & FAR$_{0.8}$ & FAR$_{0.5}$ & FAR$_{0.8}$ \\ \midrule
    CLIP-TSA \cite{CLIP-TSA} & Weak & \cellcolor[HTML]{EEEDC1}0.055 & \cellcolor[HTML]{F3EFC2}0.034 & \cellcolor[HTML]{008585}1.000 & \cellcolor[HTML]{008585}1.000 & - & - & - & - \\
    RTFM \cite{RTFM} & Weak & \cellcolor[HTML]{DBE4BC}0.129 & \cellcolor[HTML]{EFEDC1}0.049 & \cellcolor[HTML]{2D9991}0.823 & \cellcolor[HTML]{5FAE9D}0.625 & \cellcolor[HTML]{EFEDC1}0.050 & \cellcolor[HTML]{F6F0C3}0.020 & \cellcolor[HTML]{C0D9B6}0.236 & \cellcolor[HTML]{F0EDC2}0.046 \\
    PEL \cite{PEL} & Weak & \cellcolor[HTML]{FAF2C4}0.005 & \cellcolor[HTML]{FBF2C4}0.001 & \cellcolor[HTML]{6EB5A1}0.562 & \cellcolor[HTML]{99C8AC}0.393 & \cellcolor[HTML]{FBF2C4}0.001 & \cellcolor[HTML]{FBF2C4}0.000 & \cellcolor[HTML]{D7E3BB}0.145 & \cellcolor[HTML]{ECECC1}0.063 \\
    VadCLIP \cite{vadclip} & Weak & \cellcolor[HTML]{F5F0C3}0.024 & \cellcolor[HTML]{F9F2C4}0.008 & \cellcolor[HTML]{8DC2A9}0.442 & \cellcolor[HTML]{B2D3B2}0.292 & \cellcolor[HTML]{E8EAC0}0.077 & \cellcolor[HTML]{EFEDC1}0.050 & \cellcolor[HTML]{22948E}0.866 & \cellcolor[HTML]{64B19E}0.604 \\
    MGFN \cite{MGFN} & Weak & \cellcolor[HTML]{F3EFC2}0.033 & \cellcolor[HTML]{F8F1C4}0.012 & \cellcolor[HTML]{B0D2B2}0.300 & \cellcolor[HTML]{B2D2B2}0.294 & \cellcolor[HTML]{6EB5A1}0.565 & \cellcolor[HTML]{9CC9AC}0.382 & \cellcolor[HTML]{008585}1.000 & \cellcolor[HTML]{008585}1.000 \\
    UR-DMU \cite{URDMU} & Weak & \cellcolor[HTML]{F9F1C4}0.011 & \cellcolor[HTML]{FBF2C4}0.002 & \cellcolor[HTML]{CEDFB9}0.180 & \cellcolor[HTML]{D1E0BA}0.171 & \cellcolor[HTML]{D9E4BC}0.136 & \cellcolor[HTML]{EBECC0}0.064 & \cellcolor[HTML]{4CA698}0.700 & \cellcolor[HTML]{52A99A}0.676 \\
    UR-DMU \cite{GlanceVAD} & Glance & \cellcolor[HTML]{F8F1C4}0.016 & \cellcolor[HTML]{FBF2C4}0.003 & \cellcolor[HTML]{C5DBB7}0.219 & \cellcolor[HTML]{DBE4BC}0.129 & - & - & - & - \\
    UR-DMU \cite{HolmesVAU} & Full & \cellcolor[HTML]{FAF2C4}0.005 & \cellcolor[HTML]{FBF2C4}0.000 & \cellcolor[HTML]{D9E4BC}0.136 & \cellcolor[HTML]{FBF2C4}0.000 & - & - & - & - \\ \bottomrule
    \end{tabular}%
\end{table*}

\section{Experiments}

\subsection{Baselines}
\label{sec:baselines}
This paper evaluates existing models on large-scale datasets using our proposed metrics and benchmarks. To facilitate comprehensive comparison, we implement 10 SOTA VAD methods, which can be categorized into four methodological groups: 1) traditional methods (RTFM \cite{RTFM}, MGFN \cite{MGFN}, UR-DMU \cite{URDMU}) employing the common weakly-supervised paradigm without external knowledge; 2) text-enhanced methods (PEL \cite{PEL}, VadCLIP \cite{vadclip}, TEVAD \cite{TEVAD}, CLIP-TSA \cite{CLIP-TSA}) that integrate vision-language pre-trained models for performance enhancement; 3) zero-shot methods (LAVAD \cite{LAVAD}) leveraging large models' generalization capabilities to mitigate data bias and perform VAD task without tuning; and 4) methods utilizing additional supervision (GlanceVAD \cite{GlanceVAD}, HolmesVAU \cite{HolmesVAU}), which improve accuracy at the expense of requiring labor-intensive annotation.
It is worth noting that UR-DMU, GlanceVAD, and HolmesVAU use the same backbone, thus allowing us to compare the performance of different supervision paradigms more fairly.
For the experimental setup, we retrain RTFM and MGFN on UCF-Crime, retrain UR-DMU, PEL, and VadCLIP on MSAD, and directly employ official open-source weights for all other methods.

\subsection{Benchmarking Existing Methods}
We report the performance of SOTA VAD methods with our proposed probabilistic AUC/AP, LaAP and hard normal benchmarks in Tables~\ref{tab:benchmarking-1},\ref{tab:benchmarking-2}.
From Table~\ref{tab:benchmarking-1}, we observe that using probabilistic instead of original AUC/AP values significantly impacts method rankings, demonstrating that metric reliability is affected by annotator disagreements. This effect is most pronounced on UCF-Crime, while being relatively minor on XD-Violence and MSAD, which aligns with our observations of annotation discrepancy discussed in Section~\ref{sec:analysis_bias}.
Furthermore, comparative analysis of LaAP metrics reveals that models with comparable AUC/AP may exhibit substantial differences in LaAP. For instance, on UCF-Crime, MGFN underperforms other methods in both AUC and AP metrics but achieves the highest LaAP among weakly-supervised approaches. Similarly on MSAD, UR-DMU and VadCLIP demonstrate identical AUC performance yet exhibit significant LaAP discrepancy.
Experiments also confirm that methods incorporating additional supervisory signals achieve consistent improvements across all metrics. 

From Table~\ref{tab:benchmarking-2}, while most methods maintain low FAR ($\le10\%$) on the original test set, they suffer dramatic performance degradation on our proposed hard normal benchmarks. We observe a 42\% average increase at 0.5 threshold, and 36\% at 0.8 threshold, with some methods exceeding 70\% FAR. This reveals severe scene overfitting in existing approaches.
Comparative studies of supervision strategies using the same UR-DMU backbone reveal that clearer supervisory signals can mitigate overfitting risks.
Although current anomaly videos in the training sets contain both normal and abnormal intervals, video-level weak supervision struggles to guide models in learning true anomaly patterns.

\begin{figure*}[t]
  \centering
  \hfill
  \begin{minipage}[t]{0.30\textwidth}
    \centering
    \includegraphics[width=0.85\linewidth, height=4.5cm, keepaspectratio=false]{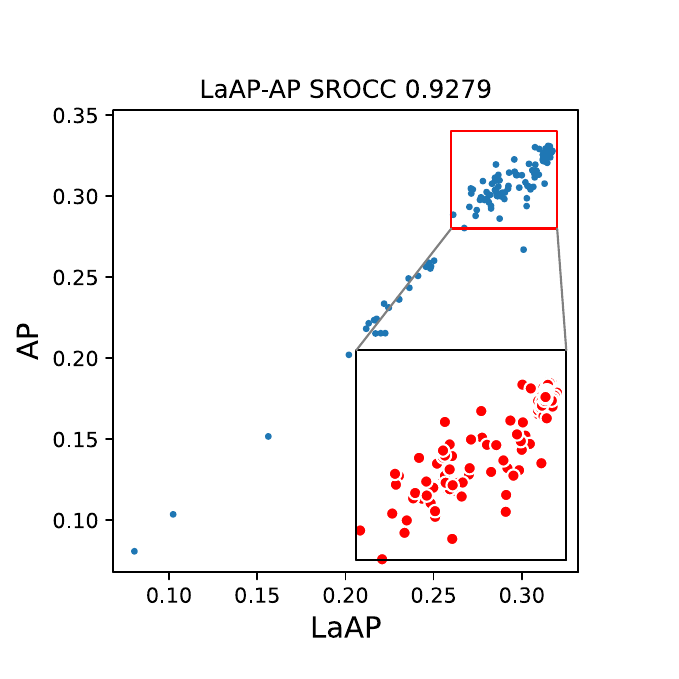}
    \caption{Consistency analysis between LaAP and AP, where each data point represents the metric values for a model at a specific training step.}
    \label{fig:exp-srocc}
  \end{minipage}
  \hfill
    \begin{minipage}[t]{0.34\textwidth}
    \centering
    \includegraphics[width=\linewidth, height=4.5cm, keepaspectratio=false]{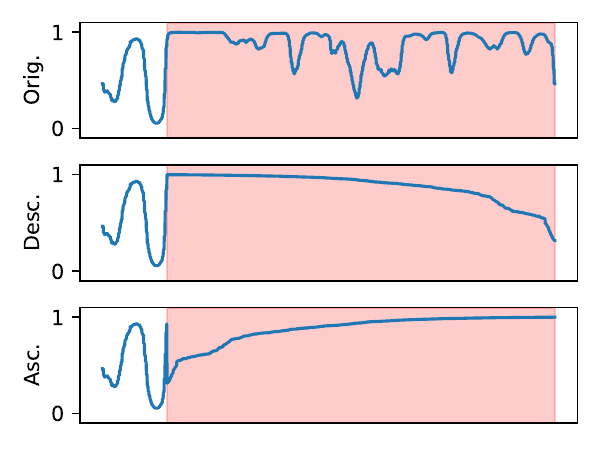}
    \caption{Visualization of different score modification methods: \textit{Desc.} re-arranges predictions within the abnormal area in descending order; \textit{Asc.} does so in ascending order.}
    \label{fig:exp-timely-mod}
  \end{minipage}
  \hfill
  \begin{minipage}[t]{0.31\textwidth}
    \centering
    \includegraphics[width=\linewidth, height=4.5cm, keepaspectratio=false]{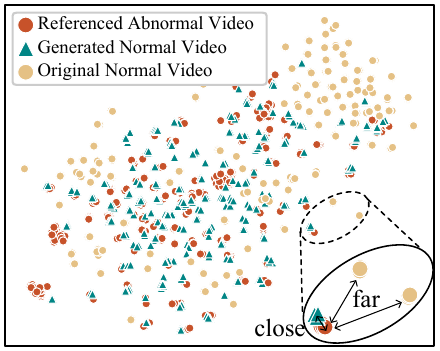}
    \caption{T-SNE visualization illustrating the distribution of reference abnormal videos, generated normal videos, and original normal videos.}
    \label{fig:exp-dist-tsne}
  \end{minipage}
  \hfill
\end{figure*}

\subsection{Reliability of LaAP}
In Figure~\ref{fig:exp-srocc}, we plot the AP and LaAP values and calculate the Spearman's Rank-Order Correlation Coefficient (SROCC) during the training of VadCLIP, which shows strong correlation between LaAP and AP.
To investigate the advantages of LaAP, we conduct a comparative analysis by modifying predicted scores from VadCLIP \cite{vadclip} and PEL \cite{PEL} on UCF-Crime \cite{ucf-crime}. As illustrated in Figure~\ref{fig:exp-timely-mod}, we simulate early/late detection through temporal reordering of anomaly scores within abnormal segments while preserving original score magnitudes. The \textit{Desc.} makes the detection earlier by sorting scores in descending order, while \textit{Asc.} achieves the opposite effect through ascending order.
Table~\ref{tab:exp-timely-mod} demonstrates LaAP's latency awareness: While AUC/AP remain invariant, our LaAP effectively rewards earlier detections through its temporal weighting mechanism.

\begin{table}[t]
  \centering
  \caption{Comparisons between AUC, AP, and LaAP under different score modification methods (visualized in Fig.~\ref{fig:exp-timely-mod}). LaAP is able to effectively reward earlier detections.}
  \label{tab:exp-timely-mod}
  \begin{tabular}{@{}lcccc@{}}
  \toprule
  Model                    & Modification & AUC   & AP    & LaAP  \\ \midrule
  \multirow{3}{*}{VadCLIP} & Desc.        & 0.900 & 0.486 & 0.783 \\
                           & Ori.         & 0.900 & 0.486 & 0.740 \\
                           & Asc.         & 0.900 & 0.486 & 0.675 \\ \midrule
  \multirow{3}{*}{PEL}     & Desc.        & 0.886 & 0.439 & 0.744 \\
                           & Ori.         & 0.886 & 0.439 & 0.727 \\
                           & Asc.         & 0.886 & 0.439 & 0.645 \\ \bottomrule
  \end{tabular}%
\end{table}

\subsection{Domain Gap of UCF-HN and MSAD-HN}
To validate the realism of our synthetic data, we compute Fréchet Inception Distance (FID) \cite{fid} and Fréchet Video Distance (FVD) \cite{fvd} between UBNormal \cite{ubnormal} and our method on real-world surveillance videos, which comprise videos of original test sets in UCF-Crime and MSAD. As in Table~\ref{tab:exp-fid}, significantly lower values demonstrate the superior realism of our benchmarks.

To further rule out the influence of the domain gap, we calculate the FAR on MSAD-HN using models trained on UCF-Crime, and compare it with results on the original UCF-Crime test set and UCF-HN. The videos in MSAD-HN are generated and contain scenes not present in the training set (i.e., UCF-Crime). If domain gap were the main cause of the observed false alarms, a high FAR on MSAD-HN would also be expected.
However, experimental results in Tab.~\ref{tab:domain_gap} demonstrate a relatively low false alarm rate on MSAD-HN, suggesting that the high false alarms are not attributable to the real-synthetic domain gap. This confirms that our benchmarks can reliably detect scene-specific overfitting.

\begin{figure*}[htbp]
   \centering
   \hfill
   \begin{subfigure}[t]{0.35\textwidth}
        \centering
        \includegraphics[width=\linewidth]{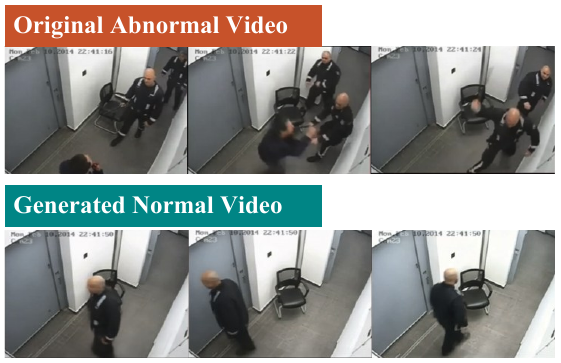}
        \subcaption{Comparison between a reference video and the corresponding generated video in UCF-HN.}
        \label{fig:exp-vis-i2v-samples}
   \end{subfigure}
   \hfill
   \begin{subfigure}[t]{0.55\textwidth}
        \centering
        \includegraphics[width=\linewidth]{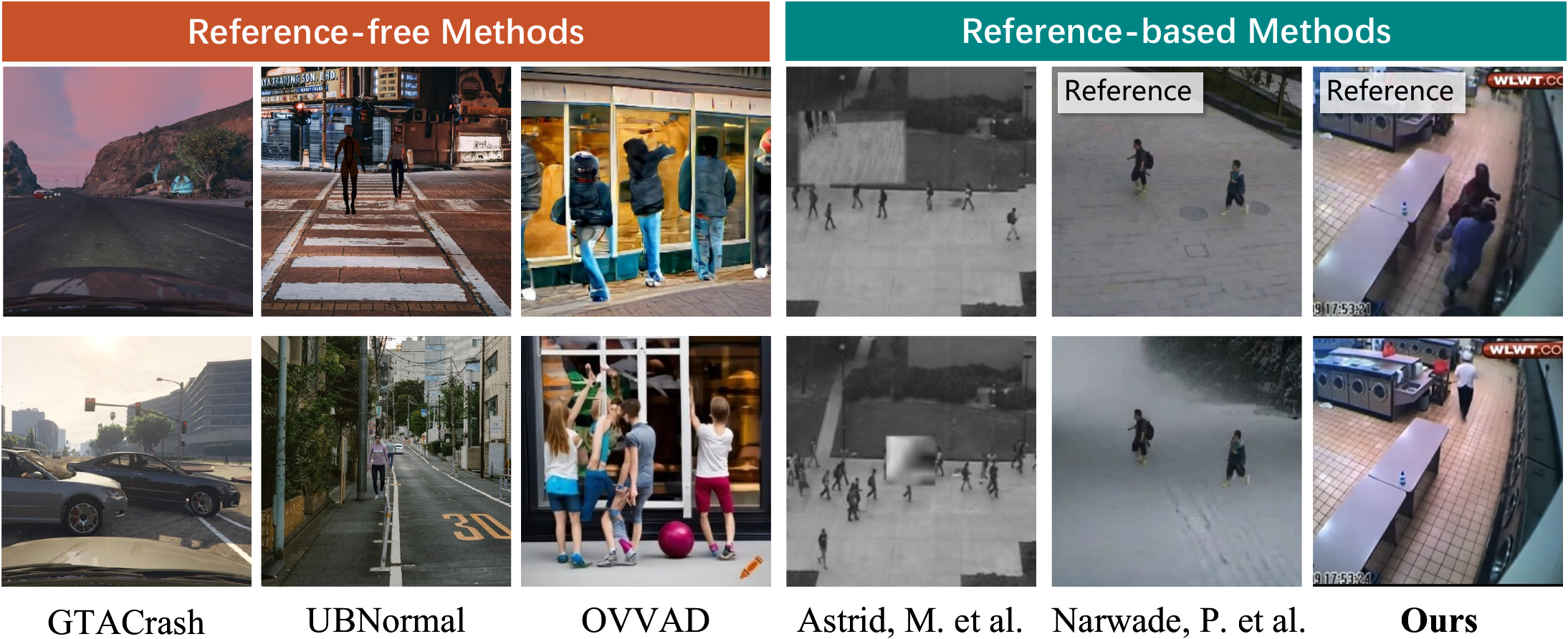}
        \subcaption{Comparisons with other VAD methods using synthetic data. Reference based methods indicate methods that synthesize videos based on existing datasets, while reference-free methods do not rely on existing datasets.}
        \label{fig:exp-vis-i2v-others}
   \end{subfigure}
   \hfill
   \caption{Comparison of synthetic video data generated by existing methods, including GTACrash \cite{GTACrash}, UBNormal \cite{ubnormal}, OVVAD \cite{OVVAD}, Astrid, M. et al. \cite{Astrid2021}, and Narwade, P. et al. \cite{NarwadeICPR24}.}
   \label{fig:exp-vis-synthetic}
\end{figure*}

\begin{table}[t]
  \centering
  \caption{Comparisons between UBNormal, UCF-HN and MSAD-HN with FID \cite{fid} and FVD \cite{fvd} metrics.}
  \label{tab:exp-fid}
    \begin{tabular}{@{}lcc@{}}
    \toprule
    Dataset & FID $\downarrow$    & FVD $\downarrow$    \\ \midrule
    UBNormal \cite{ubnormal} & 202.6 & 698.0 \\
    UCF-HN                   & 149.5 & 530.2 \\
    MSAD-HN                  & 140.4 & 461.0 \\ \bottomrule
    \end{tabular}%
\end{table}
\begin{table}[t]
\centering
\caption{Analysis of domain gap with FAR$_{0.5}$.}
\label{tab:domain_gap}
\begin{tabular}{@{}lcccc@{}}
\toprule
\multirow{2}{*}{Test-set} & \multicolumn{4}{c}{Model (trained on UCF-Crime)}       \\ \cmidrule(l){2-5} 
                          & RTFM \cite{RTFM}  & PEL \cite{PEL}   & VadCLIP \cite{vadclip} & MGFN \cite{MGFN}  \\ \midrule
UCF                       & 0.129 & 0.005 & 0.024   & 0.033 \\
UCF-HN                    & 0.823 & 0.562 & 0.422   & 0.300 \\
MSAD-HN                   & 0.282 & 0.124 & 0.201   & 0.116 \\ \bottomrule
\end{tabular}%
\end{table}

\begin{figure}[t]
    \centering
    \begin{subfigure}[b]{\linewidth}
        \includegraphics[width=\textwidth]{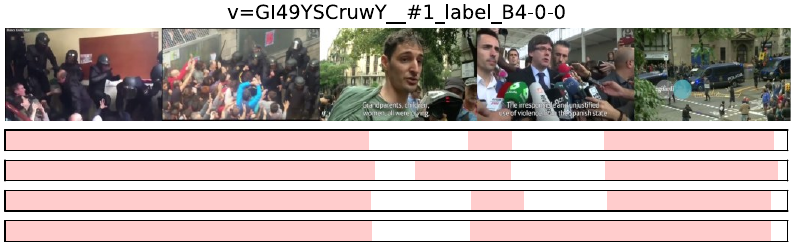}
        \caption{Different annotation granularity.}
        \label{fig:anno-vis-1}
    \end{subfigure}

    \begin{subfigure}[b]{\linewidth}
        \includegraphics[width=\textwidth]{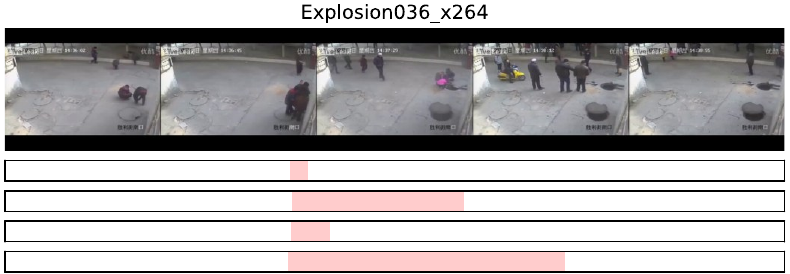}
        \caption{The start time is more consistent than the end time.}
        \label{fig:anno-vis-2}
    \end{subfigure}

    \caption{Examples of discrepancies across four rounds of annotations. Red regions indicate the annotated anomalies.}
    \label{fig:anno-vis}
\end{figure}

\begin{figure}[t]
    \centering
    \includegraphics[width=\linewidth]{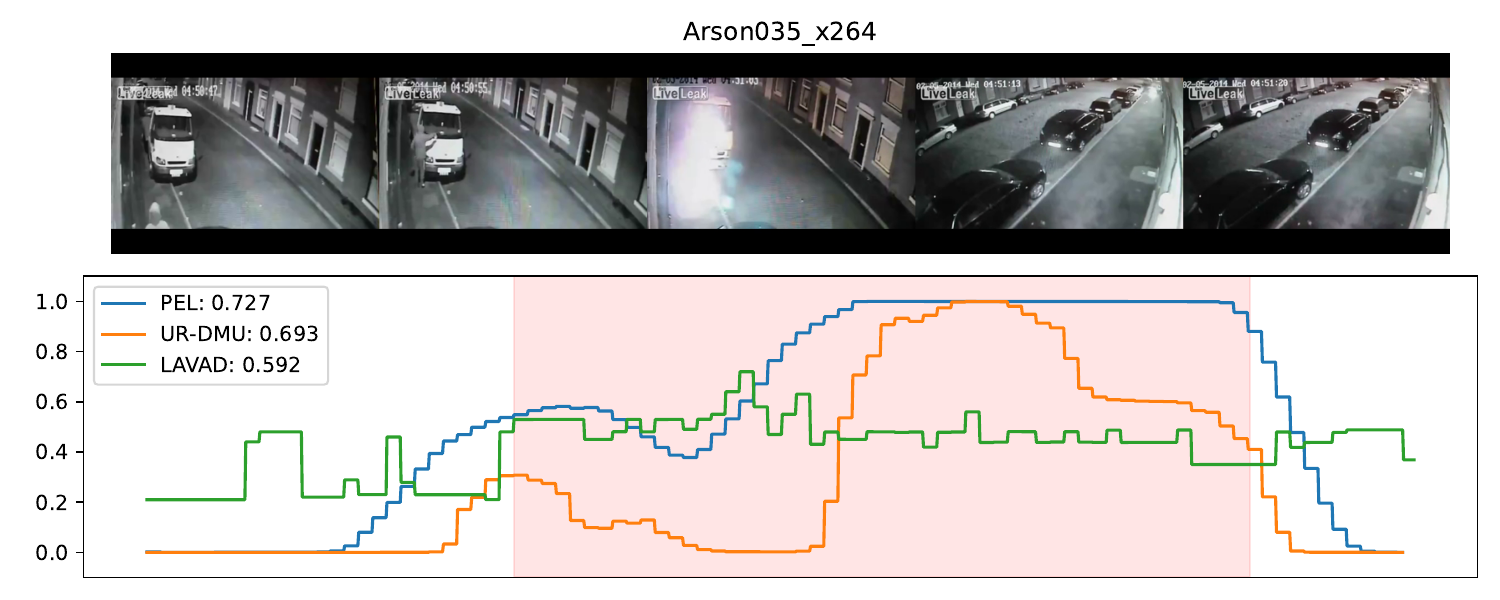}
    \caption{Example of predictions of models with different LaAP. The dataset-level LaAP for each model is listed in the legend.}
    \label{fig:laap-preds-vis-1}
\end{figure}

We also provide visualized comparisons. 
Fig.~\ref{fig:exp-dist-tsne} compares the similarity of the generated normal videos and the original normal videos with the abnormal videos in the feature space, where we use \texttt{CLIP-ViT-B/16} \cite{CLIP} as the feature extractor and T-SNE to visualize sampled frames of each video. 
Our generated normal videos (hard negative samples) have closer proximity to abnormal videos than the original normal videos, which can better evaluate scene overfitting.
In Fig.~\ref{fig:exp-vis-i2v-samples}, we compares an original anomaly video (showing ``Abuse'') with its generated normal counterpart, where violent actions are replaced by natural walking while maintaining perfect background coherence. 
Fig.~\ref{fig:exp-vis-i2v-others} further compares with existing synthetic data approaches in VAD. Notably, all compared methods differ in motivation from our work, with the comparison focused solely on generation quality. Our method features more natural dynamics and a higher degree of foreground-background integration.

\subsection{Visualization}
\begin{figure}[t]
    \centering
    \includegraphics[width=\linewidth]{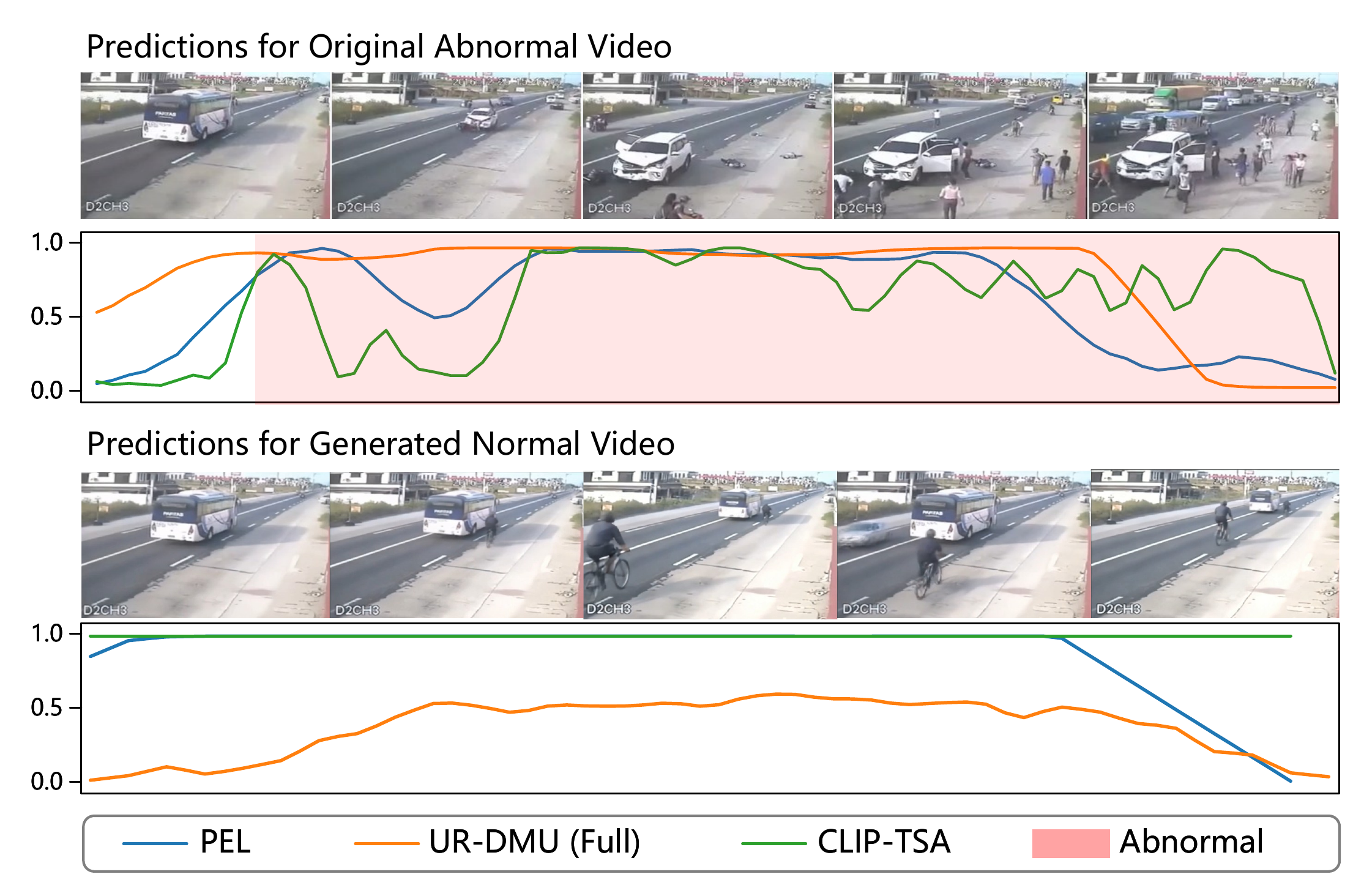}
    \caption{Examples of prediction results for the original abnormal video and its generated normal counterpart. All methods yield false alarms.}
    \label{fig:HN-preds-vis-1}
\end{figure}

Fig.~\ref{fig:anno-vis} presents two typical examples of annotation discrepancies, where they diverge in granularity and determination of ending (Finding 1).
Fig.~\ref{fig:laap-preds-vis-1} visualizes different models' predictions of a video, where PEL demonstrates a significant improvement in prediction scores within the anomaly interval when compared to UR-DMU and LAVAD. Although LAVAD exhibits higher scores during the initial phase of the anomaly interval, UR-DMU manifests a lower false alarm rate in the normal part. 
Note that LaAP, AUC, and AP do not have instance-level values due to their computational nature.
Fig.~\ref{fig:HN-preds-vis-1} illustrates the prediction results for the original abnormal video and its generated normal counterpart. All three methods correctly raise high anomaly scores during for the crash video. However, they also raise high false alarms for the normal video that shares the same scene context. This phenomenon indicates that the models overfit to scene-specific cues rather than capturing the true anomaly semantics.

\section{Conclusion}
In this paper, we reveal critical limitations in current metrics and benchmarks of VAD through systematic analysis. To address these limitations, we propose new metrics for reliable and latency-sensitive evaluation (i.e., ProbAUC, ProbAP and LaAP). We also construct new benchmarks to evaluate the model generalizability (i.e., UCF-HN and MSAD-HN). We benchmark various SOTA VAD methods according to our new evaluation methods, providing novel perspectives on evaluating VAD algorithms.

Our findings suggest that generative data provides a promising solution to the scarcity of real-world data. Nevertheless, current methods remain inadequate in capturing complex motion patterns, preserving fine details, and ensuring temporal consistency. In future work, we aim to develop a scalable pipeline for generating high-quality video anomaly datasets.

{
    \bibliographystyle{IEEEtran}
    \bibliography{main}
}

\vfill

\end{document}